\documentclass{article}
\usepackage{} 
\usepackage{nips13submit_e,times}
\usepackage{hyperref}
\usepackage{url}

\usepackage{graphicx} 
\usepackage{subfigure}
\usepackage{algorithm}
\usepackage{algorithmic}
\usepackage{multirow}
\usepackage{caption}
\usepackage{verbatim}
\newcommand{\minitab}[2][l]{\begin{tabular}{#1}#2\end{tabular}}

\title{Learning High-level Image Representation for Image Retrieval via Multi-Task DNN using Clickthrough Data}

\author{
Yalong Bai\\
Harbin Institute of Technology\\
\texttt{ylbai@mtlab.hit.edu.cn} \\
\And
Kuiyuan Yang \\
Microsoft Research \\
\texttt{kuyang@microsoft.com} \\
\And
Wei Yu \\
Harbin Institute of Technology \\
\texttt{w.yu@hit.edu.cn} \\
\And
Wei-Ying Ma \\
Microsoft Research \\
\texttt{wyma@microsoft.com} \\
\And
Tiejun Zhao \\
Harbin Institute of Technology\\
\texttt{tjzhao@hit.edu.cn} \\
}

\nipsfinalcopy 

\begin{document}

\maketitle

\begin{abstract}
Image retrieval refers to finding relevant images from an image database for a query, which is considered difficult for the gap between low-level representation of images and high-level representation of queries. Recently further developed Deep Neural Network sheds light on automatically learning high-level image representation from raw pixels. In this paper, we proposed a multi-task DNN for image retrieval, which contains two parts, i.e., query-sharing layers for image representation computation and query-specific layers for relevance estimation. The weights of multi-task DNN are learned on clickthrough data by Ring Training. Experimental results on both simulated and real dataset show the effectiveness of the proposed method.
\end{abstract}

\section{Introduction}
\begin{figure}[!ht]
\centering
\begin{minipage}{0.45\linewidth}
\includegraphics[width=1\textwidth,page=6]{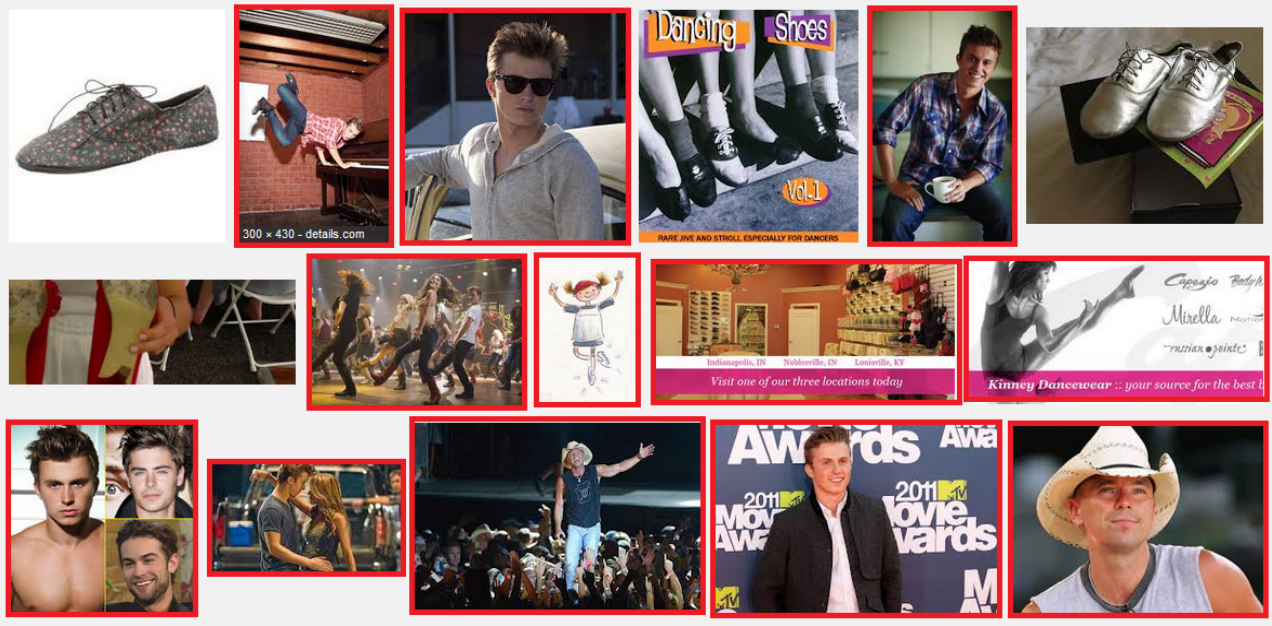}
\caption{The top ranked images for query ``kenny dance shoe" from a popular commercial image search engine at September 15th, 2013. Though the surrounding texts of these images all contain ``kenny dance shoe'', the images marked with red boxes are all irrelevant ones.}\label{example}
\end{minipage}
\hfill
\begin{minipage}{0.45\linewidth}
\includegraphics[width=1\textwidth,page=5]{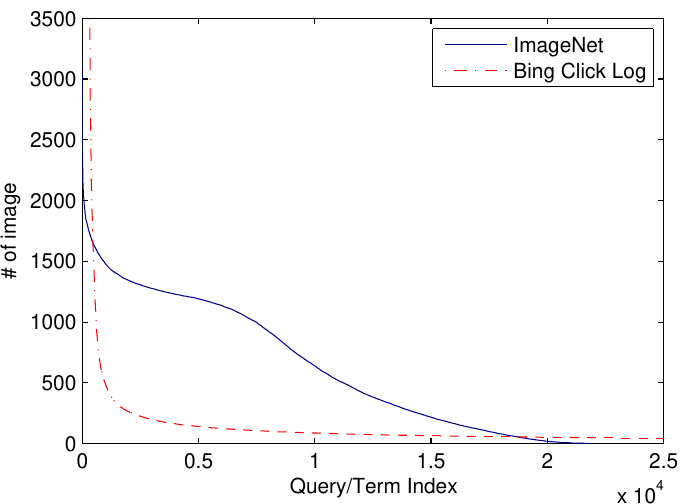}
\caption{Data distributions on ImageNet and clickthrough data.}\label{imagenet_bing_diff}
\end{minipage}
\end{figure}

Image retrieval is a challenge task in current information retrieval systems, as relevance between query (high-level semantic representation) and image (low-level visual representation) is hard to compute for the well-known semantic gap problem. In current image retrieval system, images are indirectly represented by their surrounding texts from web pages, which contain many noises and bring irrelevant images in the search results (Fig.~\ref{example} shows an example query's search results to illustrate the shortcomings of surrounding texts as representation). To improve the search results by lowering the rank of irrelevant images, binary classifier based on visual representations has been trained for each query to rerank the search results\cite{jain2011learning}. However the used visual representations such as SIFT~\cite{lowe2004distinctive}, HOG~\cite{dalal2005histograms}, and LBP~\cite{ojala2002multiresolution} are still too low-level to capture the semantic information in images~\cite{schroff2011harvesting}.

With large number of training data, convolutional deep neural network~\cite{krizhevsky2010convolutional, turaga2010convolutional, lee2009convolutional, pinto2009high} has demonstrated its great success in learning high-level image representation from raw pixels~\cite{le2012building, krizhevsky2012imagenet}, and achieved superior performance in image classification task on ImageNet~\cite{zeiler2013}. For image retrieval task, large scale clickthrough data (contains millions queries and their clicked images by users) is available as training data~\cite{craswell2007random}. The clickthrough data is different from training data for classification in the following three aspects:
\begin{enumerate}
\item The query set is much larger than category set.
\item The image number distribution on queries is significant heavy-tailed, while distributions on training data for classification (i.e., ImageNet, CIFAR-10, MINIST etc) are relatively uniform. Fig.~\ref{imagenet_bing_diff} statistics the image number distributions on ImageNet and clickthrough data from one year's search log of Bing. Compared with ImageNet, the clickthrough data is significant heavy-tailed with more than 96\% queries have less than 1000 clicked images.
\item The concept of many queries are not exclusive, e.g. ``dog" vs ``puppy".
\end{enumerate}
The three differences make exiting binary DNN and multi-class DNN not suitable as models. Binary DNN suffers from the limited training data for each query especially the large number of tail queries, while multi-class DNN cannot handle millions queries and inclusive problem between queries. To leverage the supervised information in clickthrough data, we proposed a new DNN model named as multi-task DNN. In multi-task DNN, ranking images for a query is treated as a task, while all queries share the same image representation layers. In addition, we proposed ring training which simultaneously updates the shared weights and query specific weights.

The rest of paper is structured as follows. In Section 2, we introduce multi-task DNN for image retrieval and define the objective function based on the clickthrough data. In Section 3, we introduce the Ring Training, a transfer learning mechanism for training multi-task DNN. Section 4 and Section 5 give the simulated and real experimental verification of the proposed method. Finally, we conclude the paper in Section 6.

\section {Multi-task DNN}
\begin{figure}[t]
\centering
\includegraphics[width=1\textwidth,page=9]{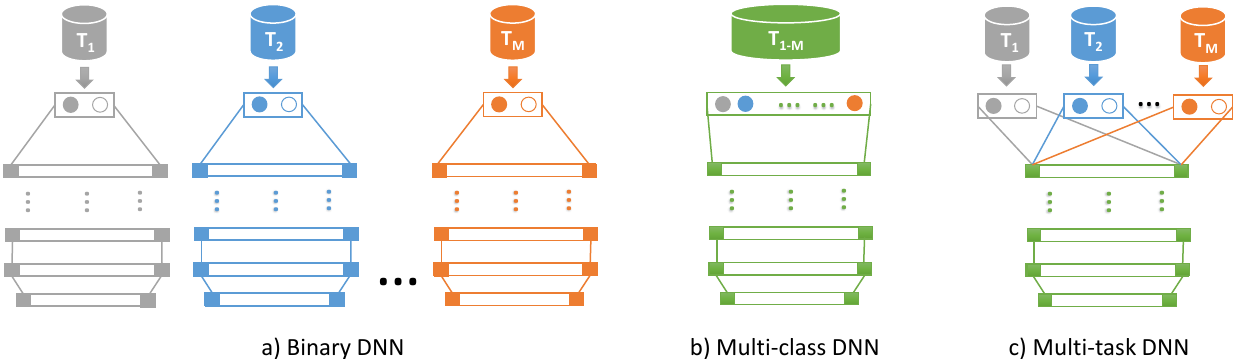}
\caption{The architectures of binary DNN, multi-class DNN and multi-task DNN.}\label{Figthreemodel}
\end{figure}

Multi-task DNN as illustrated in Fig.~\ref{Figthreemodel}(c) contains two parts: query-sharing layers and query-specific layers. Based on multi-task DNN, relevance score between query $q$ and image $I$ is defined as,

\begin{equation}
r(I, q) = \psi\big(\phi(I;W_s);W_q\big),\label{EqRankFunction}
\end{equation}
where $\phi(I;W_s)$ generates the image representation, $W_s$ are the weights of query-sharing layers, and $\psi(\cdot;W_q)$ computes the relevance, $W_q$ are the weights of query-specific layers.

Given a clickthrough dataset contains $M$ queries denoted by  $\{q_i\}_{i=1}^M$, each query $q_i$ with $n_i$ training images denoted as $\{I_j^i\}_{j=1}^{n_i}$, we define the objective function that will be used for training as following

\begin{equation}
\min_\Theta J(\Theta) = \frac{1}{N}\sum_{i=1}^{M}\sum_{j=1}^{n_i} L\big(g(I_j^i, q_i),r(I_j^i,q_i)\big),\label{EqObj}
\end{equation}

where $\Theta=\{W_s, \{W_{q_i}\}_{i=1}^{M}\}$ denotes all the weights in the model, $N=\sum_{i=1}^M n_i$ is the total number of training images, $g(I_j^i,q_i)\in\{-1,+1\}$ is the groundtruth denotes whether image $I_i^j$ is clicked by query $q_i$, $r(I_j^i,q_i)$  is the ranking score defined in Eq.~\ref{EqRankFunction}, $L\big(g(I_j^i, q_i),r(I_j^i,q_i)\big)$ is the loss function penalizes the inconsistence between groundtruth and prediction.

To optimize the objective function, we resort to gradient descent method. The gradient of the objective function with respect to $W_{q_i}$ is
\begin{equation}
\nabla_{W_{q_i}}J = \frac{1}{N}\sum_{j=1}^{n_i}\frac{\partial L\big(g(I_j^i, q_i),r(I_j^i,q_i)\big)}{\partial W_{q_i}},
\end{equation}
which only need to average the gradients over training images of query $q_i$.
The gradient of the objective function with respect to $W_s$ is
\begin{equation}
\nabla_{W_{s}}J = \frac{1}{N}\sum_{i=1}^{M}\sum_{j=1}^{n_i}\frac{\partial L\big(g(I_j^i, q_i),r(I_j^i,q_i)\big)}{\partial W_s},
\end{equation}
where gradients of all training images from all queries are averaged. Computing gradient in batch mode is computational intensive over large-scale clickthrough data, which is also computational inefficient as many queries share similar concept (As an extreme example, the dataset only contains two queries ``cat'' and ``kitten'', and the training images for each query are exactly the same, updates the weights iteratively using average gradients of ``cat'' and ``kitten'' will be two times faster than batch mode, which shares similar advantage as mini-batch mode).

\section {Ring Training}
\begin{figure}[ht]
\centering
\includegraphics[width=0.65\textwidth,page=8]{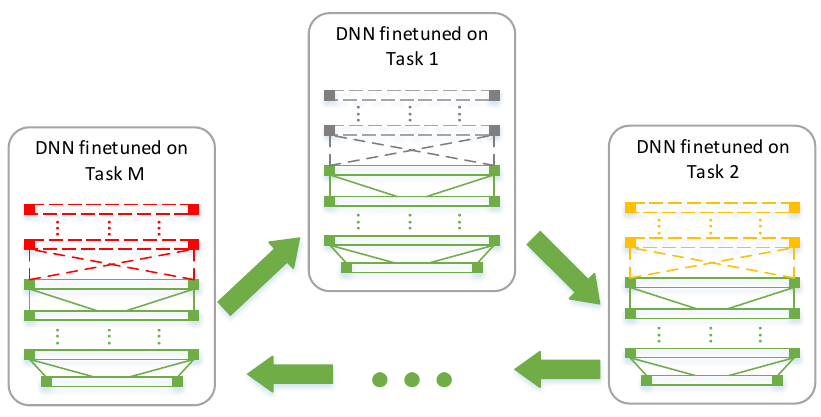}
\caption{Schematic illustration of Ring Training. Green layers are shared between all queries, and layers of other colors are query specific layers. Ring training loops over the queries to update the weights.}\label{FigRing}
\end{figure}

Based on the above observation, we proposed ring training to update the weights as illustrated in Fig.~\ref{FigRing}. Ring training loops several rounds over queries, each query $q_i$ updates both $W_{q_i}$ and $W_s$ several epoches with the average gradients of the query's training images $\{I_{ij}\}_{j=1}^{n_i}$ in batch or mini-batch mode. Ring training shares similar advantage as mini-batch mode and ensures faster convergence rate. From the viewpoint of transfer learning, ring training transferred the image representation from previously learned queries to current query,  and can avoid overfitting even the query with few training images. The detailed algorithm is summarized in Algorithm 1. In practice, the learning rate $\eta_s$ for $W_s$ is gradually reduced to 0 several rounds before $\eta_q$, after $\eta_s$ reduced to 0, the layers related to image representation are fixed.


\begin{algorithm}[tb]
   \caption{Procedure of Ring Training For multi-task DNN}
   \label{alg:example}
\begin{algorithmic}
   \STATE {\bfseries Input:} $M$ queries, traning images $\{I_j^i\}_{j=1}^{n_i}$ for query $q_i$, shared weights $W_s$, query specific weights $\{W_{q_i}\}_{i=1}^M$, $R$ rounds, $E$ epochs each round, learning rate $\eta_q$ and $\eta_s$
   \STATE Initialize $W_s$ randomly.
   \FOR{$r=1$ {\bfseries to} $R$}
   \FOR{$i=1$ {\bfseries to} $M$}
   \FOR{$e=1$ {\bfseries to} $E$}
   \STATE {\bfseries Forward Pass:}
   \STATE Extract public feature: $v\leftarrow \phi(I_j^i;W_s)$
   \STATE Compute output: $r(I_j^i,q_i)=\psi(v;W_{q_i})$
   \STATE {\bfseries Backpropagate Gradients:}
   \STATE Compute gradient ${A}'_{W_{q_i}}$ and ${A}'_{W_s}$ with respect to parameters $W_{q_i}$ and $W_s$:
   \STATE Update category special layers: $W_{q_i} = W_{q_i} - \eta_q {A}'_{W_{q_i}}$
   \STATE Update shared layers: $W_s = W_s - \eta_s {A}'_{W_s}$
   \ENDFOR
   \ENDFOR
   \ENDFOR
\end{algorithmic}
\end{algorithm}



\section {Experiments on CIFAR-10}
In this section, we did several simulations based on a small dataset CIFRA-10 to verify the effectiveness of the proposed method. CIFAR-10 is a dataset composed of 10 classes of natural objects, each class contains 5,000 training images and 1,000 testing images. Images are with small size of $32\times32$.

\subsection {Overview}

The architecture of network for CIFRA-10 contains four layers in order, three convolutional layers and one fully-connected layer, the first two convolutional layers have 32 filters with size of $5\times5$, the last convolutional layer has 64 filters with size of $5\times5$. The first convolutional layer is followed by a max-pooling layer, and the other two convolutional layers are followed by average-pooling layer, overlap pooling is used in pooling layer with window size of $3\times3$ and stride of 2(i.e., neighboring pooling windows overlap by 1 element). The defined network achieves 78.2\% accuracy on standard CIFAR-10 task, which is comparable to work of \cite{wan2013regularization} 78.9\% without using dropout and dropconnect. 

All experiments are based on mini-batch SGD (stochastic gradient descent) with batch size of 128 images, the momentum is fixed at 0.9, and weight decay is set as 0.004. The update rule for weights is defined as

$~~~~~~~v_{t+1} = 0.9v_t - 0.004\cdot \epsilon \cdot w_t + \epsilon \cdot g_t$

$~~~~~~~w_{t+1} = w_t + v_{t+1}$

where $g_t$ is the gradient of the cost function with respect to that parameter averaged over the mini-batch, $t$ is the iteration index and $\epsilon$ is the learning rate of weights and biases are initialized with 0.001 and 0.002 respectively.

\subsection {CIFAR-10}
To simulate the heavy-tail distribution of real dataset, we construct a dataset denoted as $dataset_1$ by sampling different amounts of images for each class. The number of training images for the ten categories of $dataset_1$ is: [5000, 4000, 3000, 2000, 1000, 500, 400, 300, 200, 100]. Considering there are categories with similar concept in real dataset, we construct another datasets $dataset_2$ by randomly splitting images of ``cat'' in $dataset_1$ into two parts named as ``cat'' and ``kitten''. For comparison, $dataset_3$ is constructed with same total number of training images as $dataset_1$ by randomly sampling 1650 images per category. For each category, negative examples are randomly selected from the other categories with the same size as positive examples. Before feeding images to the neural network, we subtracted the per-pixel mean computed over the training set for each category~\cite{hinton2012improving}.

The following three methodsare compared on the three datasets:
\begin{enumerate}
  \item Binary DNN, a separate DNN is trained for each category
  \item Multi-class DNN
  \item Multi-task DNN with ring training, the proposed method
\end{enumerate}

The results are summarized in Table~\ref{CIFAR}. In general, binary DNN performs consistently worse for the severe overfitting problem.

\begin{figure}[h]
\begin{minipage}{0.48\linewidth}
\includegraphics[width=1\textwidth,page=2]{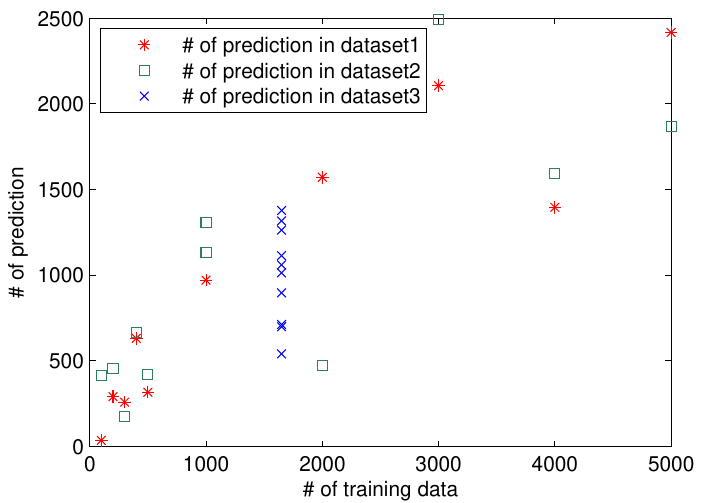}
\caption{Amount of the predicted images for each category in $dataset_1$, $dataset_2$ and $dataset_3$, using multi-class architecture}\label{predict_amount}
\end{minipage}
\hfill
\begin{minipage}{0.48\linewidth}
\includegraphics[width=1\textwidth,page=3]{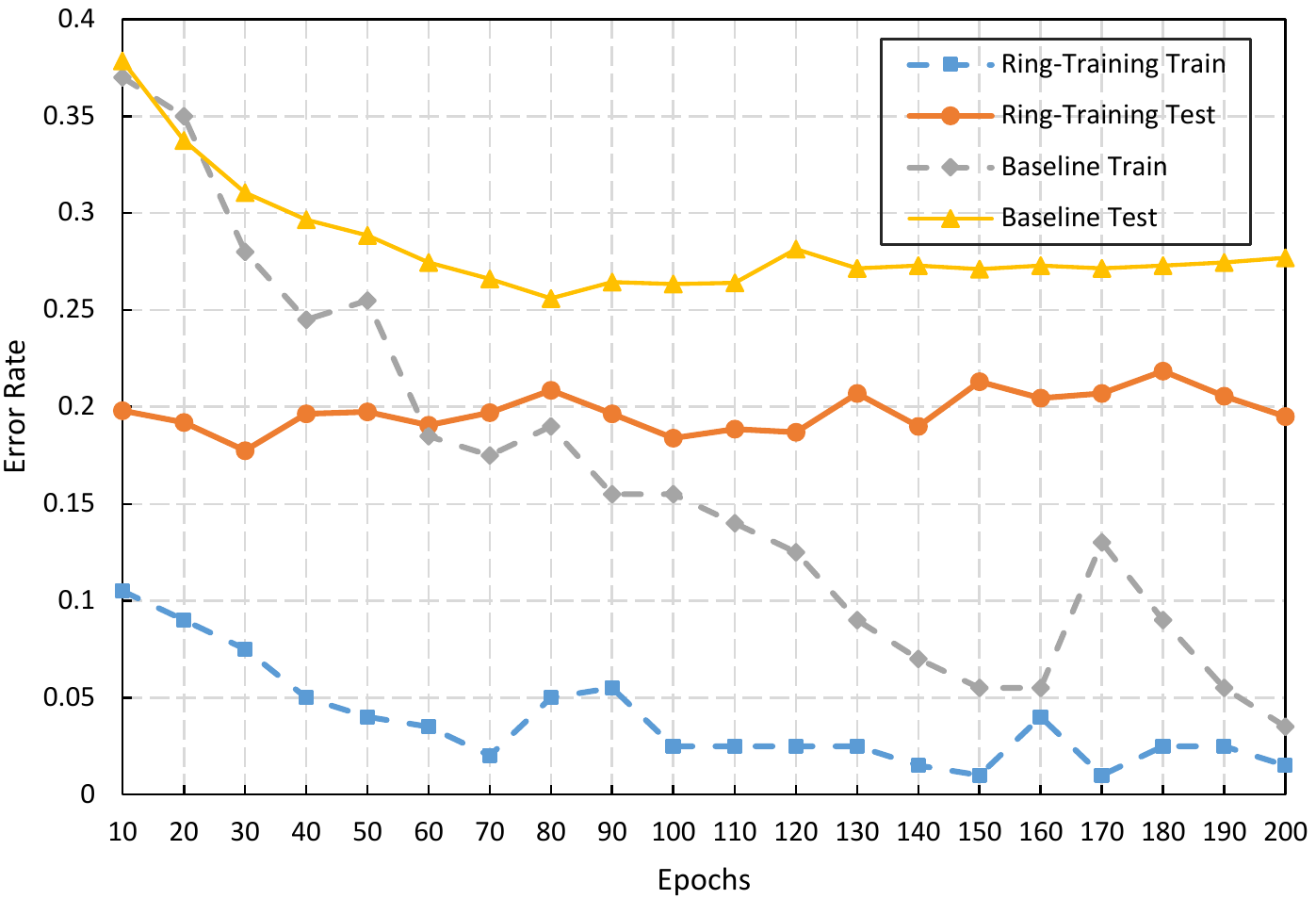}
\caption{Training progression of baseline (binary DNN) and multi-task DNN with ring training for a tail categorie.}\label{ring_progreesion}
\end{minipage}
\end{figure}

Comparing error rates on $dataset_1$ and $dataset_3$, multi-class DNN performs worse when the dataset is with heavy-tailed distribution.The performance of multi-class DNN is ever worse by comparing error rates on $dataset_1$ and $dataset_2$, where a class named cat is split into two similar class called cat and kitten~\footnote{For fair comparison, images predicted as cat and kitten are all treated as cat during evaluation.}. This demonstrates that trying to discriminate similar categories in multi-class DNN will hurt the performance. The reason is that multi-class DNN is designed to discriminate all categories, while trying to discriminate categories describing the same concept will lead overfitting problem. $dataset_1$ and $dataset_3$ are with the same number of images but with much lower test error, it is the nonuniform distribution affects the learning of multi-class DNN. Fig.~\ref{predict_amount} shows the number of predicted images vs the number of training images, top categories are overemphasized and tend to have more predicted images. In summary, all of above experiments show multi-class DNN is not suitable for real dataset, especially in image retrieval task, there is no requirement to discriminate a query from all the others, where only the relevance between image and queries is required.

In general, the proposed method achieves the lowest error rate except $dataset_3$ which is not real case. Multi-task DNN with ring training significantly outperforms the binary DNN and multi-class DNN on all nonuniform distributed datasets. Additionally, to verify how ring training improves classification error of tail categories, Fig.~\ref{ring_progreesion} shows the convergence property of a category with only 100 images in $dataset_1$. Comparing to binary DNN, multi-task DNN with ring training converges much faster (test error is table after ten epochs) and with much lower test error, which further verifys the efficiency and effectiveness of the ring training.

\begin{table*}[t]
\caption{Train and test set misclassification rate for binary DNN(separately trained), multi-class DNN and multi-task DNN.}
\label{sample-table}
\begin{center}
\begin{tabular}{c||c||c|c}
\hline
\minitab[l]{Dataset}& \minitab[l]{Model} & \minitab[l]{Train Error \%} & \minitab[l]{Test Error \% }\\
\hline
\multirow{4}*{$dateset_1$}& \minitab[l]{Binary DNN } & 31.19 & 43.27\\\cline{2-4}
& \minitab[l]{Multi-class DNN}& 6.22 & 49.82 \\\cline{2-4}
& \minitab[l]{Mutli-task DNN + ring training}& 32.79 & \textbf{39.16} \\\cline{2-4}
\hline
\hline
\multirow{4}*{$dateset_2$}& \minitab[l]{Binary DNN } & 31.09 & 43.53 \\\cline{2-4}
& \minitab[l]{Multi-class DNN}& 10.80 & 52.53 \\\cline{2-4}
& \minitab[l]{Mutli-task DNN + ring training}& 32.36 & \textbf{39.89} \\\cline{2-4}
\hline
\hline
\multirow{4}*{$dateset_3$}& \minitab[l]{Binary DNN }  & 38.79 & 43.4\\\cline{2-4}
& \minitab[l]{Multi-class DNN}& 10.41 & \textbf{31.97} \\\cline{2-4}
& \minitab[l]{Mutli-task DNN + ring training }& 30.4 & 36.68 \\\cline{2-4}
\hline
\end{tabular}
\end{center}
\label{CIFAR}
\end{table*}

\section{Experiment on image retrieval}
In this section, we verify the effectiveness of multi-task DNN in the real image retrieval task.
\subsection{Clickthrough dataset}
The clickthrough dataset, which contains 11 million queries and 1 million images and 21 million click pairs, collected from one year's search log of Bing image search, is publicly available from MSR-Bing Image Retrieval Challenge~\cite{BingGrand}, and the data distributed is same with the ``Bing Click Log" in Fig.~\ref{imagenet_bing_diff}. The dev set contains 1000 queries and 79,655 images, and the relevance between image and query are manually judged. The judgment guidelines and procedure are established to ensure high data quality and consistency.

Multi-class DNN is infeasible for such large number of queries. In this experiment, multi-task DNN with ring training is used to learn weights based on the clickthrough data.

\subsection{Experimental Setting}

The network is with five convolutional layers and two fully-connected layers, drop out with rate 0.5 is added to the first fully-connected layers during training for avoiding overfitting~\cite{hinton2012improving}. The output of the last fully-connected layer is fed to softmax to represent the relevance degree between image and query. Our network maximized the average log-probability of correct judgment about whether the image related to query. The first, second and fifth convolutional layers are followed by max-pooling layers, while the first and second max-pooling layers are followed by response-normalization layers. To accelerate the learning of early stage, the ReLU~\cite{glorot2011deep} non-linearity is applied as activation function in every convolutional and full-connected layers.

The input image is with size of $224\times224$. The first convolutional layer is with 96 filters with size of $11\times11$ and stride of 4. The second convolutional layer is with 256 filters with size of $5\times5$. The third, fourth and fifth convolutional layers are with 384 filters with size of $3\times3$. The first fully-connected layers following the fifth convolutional layer are with 4096 neurons. Three max-pooling layers are with window size of $3\times 3$ and stride of 2.

\subsection{Experimental Results}
Discounted Cumulated Gain (DCG) is adopt to as the performance metric for a ranking list. Given a ranking list for a query, the DCG is calculated as
$DCG_{25} = 0.01757 \sum_{i=1}^{25}\frac{2^{rel_i}-1}{log_2(i+1)}$
where $rel_i = {Excellent = 3, Good = 2, Bad = 0}$ is the manually judged relevance for each image with respect to the query, and 0.01757 is a normalizer to make the DCG score of 25 Excellent images to be 1. The final metric is the average of $DCG_{25}$ over all test queries.

\begin{figure*}[ht]
\centering
\includegraphics[width=1\textwidth,page=10]{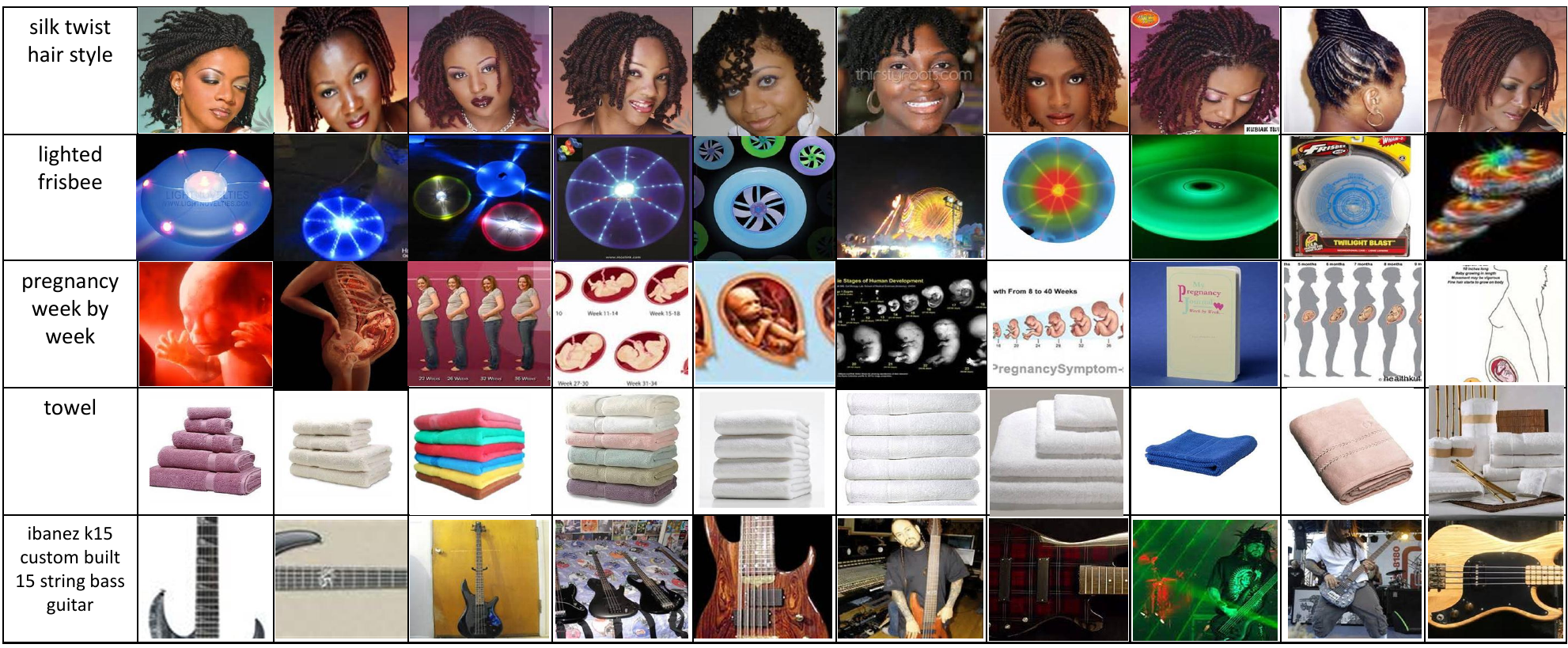}
\caption{Five randomly chose example queries with their top ranked images, the ranking of each image is computed by SVM based on learned feature.}\label{four_results}
\end{figure*}

For ring training, five convolutional layers and the first fully-connected layer are shared among all queries, the second fully-connected layer and 2-way softmax layer are used as query-specific layers. After finishing ring training, weights of sharing layers are fixed, outputs of the first fully-connected layer are used as feature for each images, then SVM to used to learn the relevance between image and query based on the extracted feature.

We compared the following three ranking methods,
\\1) Random ranker, images are randomly ranked for each query
\\2) SVM based on bag of visual words, which preselect SIFT~\cite{lowe2004distinctive} as visual feature.
\\3) SVM based on learned feature.

The results are summarized in Table~\ref{DCG}, where the learned feature achieved the best performance. Fig.~\ref{four_results} shows the ranking results of five queries based on feature learned by multi-task DNN with ring training.

\begin{table}[t]
\caption{Rank result comparison between the three ranker method}
\label{sample-table}
\begin{center}
\begin{tabular}{c||c}
\hline
Method & $DCG_{25}$ of all queries \\
\hline
Random ranker & 0.468 \\
\hline
SVM based on bag of word & 0.484 \\
\hline
SVM based on learned feature & 0.502 \\
\hline
\end{tabular}
\end{center}
\label{DCG}
\end{table}

\section{Discuss and Conclusion}

In this work, multi-task DNN learned by ring training is proposed for image retrieval. The model treats each query as a specific task, and exploits the commonalities between different tasks for image representation learning. Experimental results on both CIFAR-10 and MSR-Bing Image Retrieval Challenge show the improvement by the proposed method.

\bibliography{RT_iclr2014}

\end{document}